\documentclass{article}
\usepackage{iclr2026_conference,times}
\iclrfinalcopy 

\usepackage{amsmath,amsfonts,bm}

\def\eqref#1{equation~\ref{#1}}

\def\1{\bm{1}}

\DeclareMathAlphabet{\mathsfit}{\encodingdefault}{\sfdefault}{m}{sl}
\SetMathAlphabet{\mathsfit}{bold}{\encodingdefault}{\sfdefault}{bx}{n}

\usepackage{hyperref}
\usepackage{url}
\usepackage{graphicx}
\usepackage{booktabs}
\usepackage{amsmath}
\usepackage{amssymb}
\usepackage{xcolor}
\usepackage{caption}
\usepackage{wrapfig}
\usepackage{algorithm}
\usepackage{algpseudocode}
\title{Brand-as-Memory: Vision-Language Models Encode\\
Causal, Mechanistically Localizable Credibility Priors for News Sources}

\author{Chih-Ting Liao\thanks{Corresponding author, first author, and project lead. \texttt{mill.liao@unsw.edu.au}} \quad Xin Cao \\
University of New South Wales}

\newcommand{\dd}[1]{\texttt{#1}}

\begin{document}
\maketitle
\lhead{Preprint}

\begin{abstract}
Vision-language models (VLMs) increasingly read news and web content as images, where the publisher's identity is visually present. We show that VLMs carry a strong \emph{source-credibility prior} keyed on outlet identity, and study it along three axes. \textbf{(i)~A cross-model benchmark.} We introduce \emph{CueTrust}, a cross-model diagnostic that measures which surface source cue overrides an article's content evidence, via a Source-Override Index (SOI). Across seven VLMs and five cues the vulnerability profile is model- and scale-dependent, and the override is \emph{outlet-identity-specific and encoding-invariant}, firing from the masthead name, the logo image, or the bare domain, but not from a named author, in-text authority, or page layout (clean negative controls). \textbf{(ii)~A mechanistic account.} For the brand cue we give a full mechanistic account: swapping only the masthead moves credibility across an $\sim$11-log-odds range that tracks professional ratings ($\rho{=}0.88$ vs Media Bias/Fact Check); the prior is \emph{dual-coded} (name and logo), strengthens with scale, is \emph{causally formed at layers 19--21}, carried by interpretable seed-stable sparse-autoencoder features, and recurs at the same relative locus in a second family; it \emph{overrides} content ($\sim$1.8$\times$) as a signal-magnitude effect within a \emph{shared} pathway, not a privileged route. Steering the localized direction selectively reduces the override ($-41\%$) and generalizes to held-out outlets, confirming the prior is causally \emph{used}, not merely decodable. Deployed VLMs may thus defer to source identity over the evidence in front of them, a reliability failure we can measure across models, localize, and causally probe. We release the stimulus suite and CueTrust.
\end{abstract}

\section{Introduction}
VLMs now mediate how people read information presented as images: screenshots of news articles, social posts, and web pages. When such a model judges whether a story is credible, does its judgment turn on the \emph{evidence} in the article, or on \emph{who} the article appears to come from? We show that for current VLMs it turns substantially on the latter, that the deciding cue is the publisher's \emph{brand identity}, and (unlike prior behavioral studies) we localize where in the network that judgment is formed and what it is made of.

Concretely, we render neutral, topic-matched news articles as screenshots and vary only the masthead. A VLM's internal credibility log-odds then span an $\sim$11-log-odds range ordered almost exactly as humans rank these outlets. We call this a \emph{brand-as-memory} prior: a learned association between a source's visual identity and its perceived trustworthiness, retrieved from the model's parametric memory at the moment of judgment. Crucially the cue is \emph{visual}: a rendered brand logo, with no readable brand text, already recovers the ordering.

\begin{figure}[t]
\centering
\includegraphics[width=\linewidth]{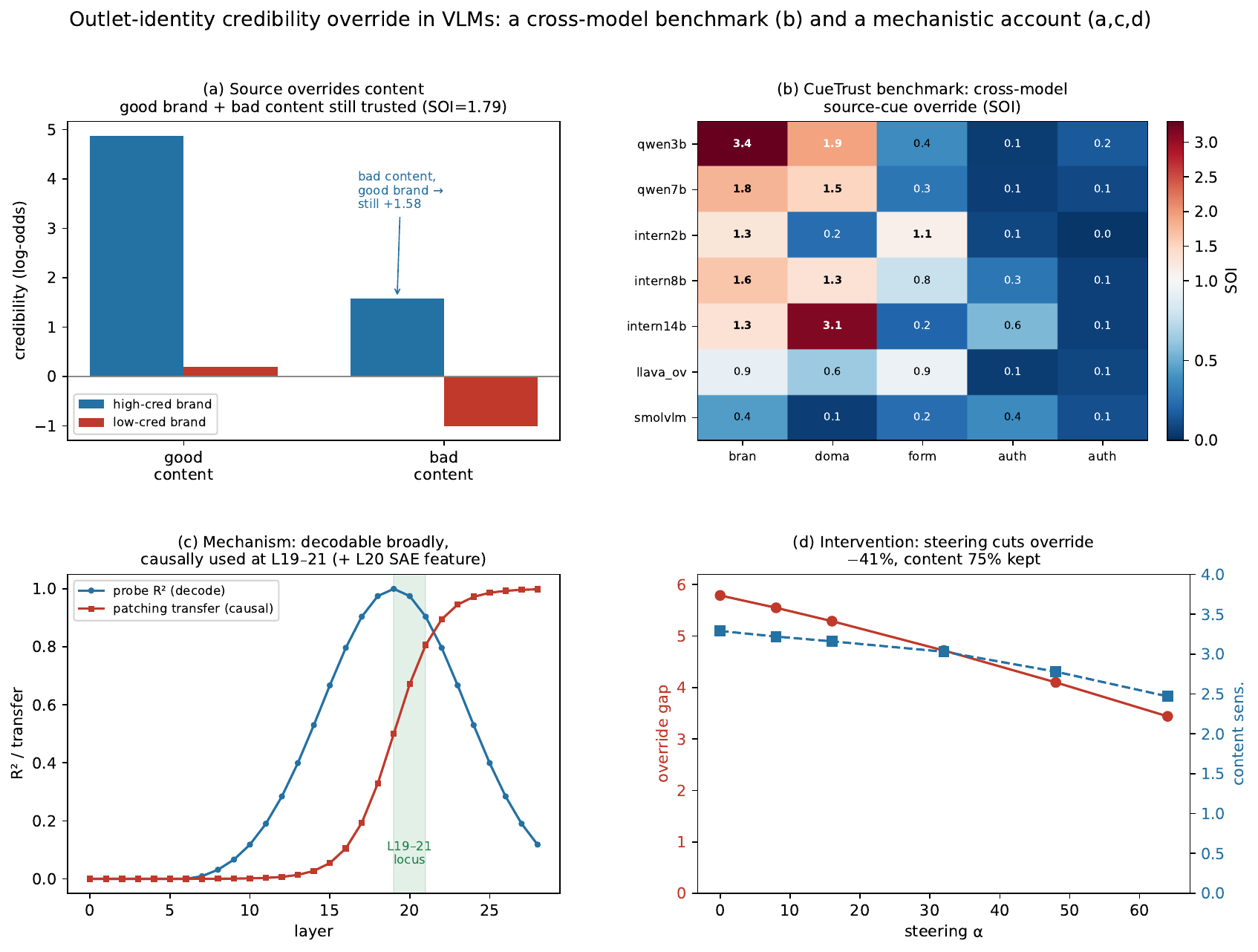}
\caption{\textbf{Overview.} VLMs carry a source-credibility prior keyed on the publishing outlet's identity, studied along three axes. (a)~\emph{Mechanism}: the prior overrides article content, a high-credibility brand with red-flag content is still judged credible (conflict cell $+1.58$; source-over-content $\mathrm{SOI}{=}1.79$), and only outlet-identity cues do so. (b)~\emph{Benchmark} (CueTrust): across seven models and five cues, brand and domain override content while author, authority, and layout do not (negative controls). (c)~\emph{Mechanism}: credibility is decodable by a linear probe peaking at L19 and causally carried through L19--21 (patching transfer $\to1$). (d)~\emph{Causal control}: steering the localized direction selectively reduces the override ($-41\%$), confirming the prior is causally used.}
\label{fig:main}
\end{figure}

We make two contributions. \textbf{(1)~A cross-model benchmark:} we introduce \emph{CueTrust}, a diagnostic measuring which surface source cue overrides an article's content via the Source-Override Index over seven models and five cues, and establish that the override is \emph{outlet-identity-specific and encoding-invariant}, triggered by the masthead name, logo, and domain but not by author, in-text authority, or layout (negative controls that bound the phenomenon), with model- and scale-dependent strength. \textbf{(2)~A mechanistic account:} we show the brand prior is continuous, dual-coded, and scale-strengthening, is causally formed at layers~19--21, carried by interpretable seed-stable sparse features, recurs in a second family, and overrides content as a shared-pathway signal-magnitude effect; steering the localized direction selectively reduces the override and generalizes to held-out outlets, confirming the prior is causally \emph{used}, not merely decodable (a debiasing prompt suppresses more but is black-box). We additionally release a controlled stimulus suite and CueTrust for source-vs-content study beyond this paper.

\smallskip\noindent\textbf{Positioning.} Concurrent text-only work \citep{khan2026agents,yang2025accuracy} shows that LLMs harbor latent source preferences that can outweigh content and persist under prompting. Our contribution is orthogonal on the two axes those studies do not touch: the phenomenon is \emph{visual} (it fires from a rendered logo image, not only an attributed name string), and we provide a \emph{mechanistic} account (localization, sparse features, a cross-family locus, and a pathway-level explanation of the override) rather than a behavioral one. Bridging a social-science phenomenon (source credibility) to a full mechanistic account in VLMs is, to our knowledge, new.

\section{Related Work}
\smallskip\noindent\textbf{Source and credibility bias in (V)LMs.}
\citet{yang2025accuracy} audit whether LLMs can rate news-source credibility from source names in isolation, finding only moderate alignment with human experts and a popularity-dependent willingness to rate at all. Most directly, \citet{khan2026agents} show across twelve LLMs that models exhibit systematic \emph{latent source preferences} based on brand identity and credentials, that these preferences can outweigh content and survive explicit instructions to ignore them, and that they grow with model scale. Our behavioral findings (\S\ref{sec:phenom},\S\ref{sec:override}) confirm this picture \emph{in VLMs}; our advance is twofold: (i) the cue is carried by a \emph{visual} brand channel (a rendered logo image alone reproduces the ordering, a modality these text-only studies cannot probe) and (ii) we supply the \emph{mechanism}, localizing where the judgment is formed, identifying the features that carry it, and explaining the content override at the level of pathways rather than behavior. Behavioral work on multimodal credibility and on image-presence effects on sharing \citep{ucl2025image} likewise stops short of localization, and source deference is a close cousin of sycophancy (deferring to a salient cue over the evidence \citep{sharma2023sycophancy}) and is distinct from perceptual visual-vs-knowledge conflict \citep{ortu2025seeing}, which concerns what the image \emph{shows} rather than who it is \emph{from}.

\smallskip\noindent\textbf{Mechanistic interpretability tooling.}
We build on sparse-feature circuit analysis \citep{marks2025sparse}, dictionary-learning and scaled SAEs \citep{bricken2023monosemanticity,templeton2024scaling,cunningham2023sparse,gao2024scaling,lieberum2024gemmascope}, the linear-representation view of concept directions \citep{park2024linear}, activation patching and circuit analysis \citep{zhang2024patching,wang2023ioi,conmy2023acdc,vig2020causal}, control-task probing \citep{hewitt2019control,belinkov2022probing}, lens-style readouts \citep{nostalgebraist2020logitlens,belrose2023tunedlens}, and recall-style factual associations \citep{meng2022rome}; against this backdrop \citep{sharkey2025openproblems}, our brand$\rightarrow$credibility prior is a visual, behaviorally-read instance using no weight editing. In the VLM setting, \citet{nikankin2025sametask} find that modality-specific circuits implement similar functions but align only toward later layers (consistent both with our late-layer causal locus and with our finding that the name-string and logo-image channels are separable) and \citet{yang2026circuittracing} report visual/semantic features emerging around layer~20, corroborating our L19--21 locus. Broader vision and VLM interpretability (multimodal neurons \citep{goh2021multimodal}, {CLIP} decomposition \citep{radford2021clip,gandelsman2024clip}, SAEs for vision models \citep{stevens2025vision}, instruction-tuned VLMs \citep{liu2023llava}) has targeted object and concept features, not a social property like source credibility. We extend this tooling from known/unknown-entity directions to a \emph{continuous}, externally-validated credibility scale carried by a visual channel.

\smallskip\noindent\textbf{Interpreting our SAE evidence, and not debiasing.}
Recent work cautions that SAE interpretability does not by itself imply utility \citep{interp2025utility} and that single SAEs can be unstable across seeds, different initialisations recover different features \citep{paulo2025different,gadgil2025ensembling}; we therefore bind features to \emph{causal} effect via signed steering and report 5-seed stability rather than feature existence alone \citep{arad2025saes}. Finally, unlike debiasing methods \citep{debiaslens2026}, we do not propose to remove the prior: most of it is warranted, and our object of study is its mechanism and its specific failure mode.

\section{Setup and Stimuli}\label{sec:setup}
\begin{figure}[t]
\centering
\includegraphics[width=0.98\linewidth]{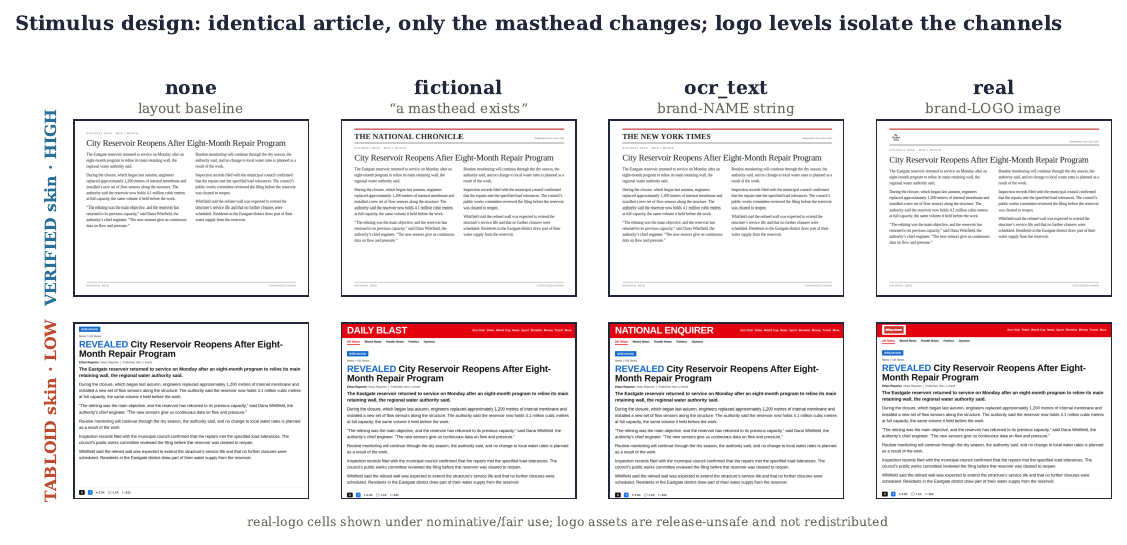}
\caption{Stimulus design (real renders). Each image is a SKIN\,$\times$\,LOGO-LEVEL composition; across a row the article is identical and only the masthead changes. The four logo levels isolate distinct cues: a layout-only baseline (\dd{none}), \emph{a masthead exists} (\dd{fictional}), the brand-name string (\dd{ocr\_text}), and the brand-logo image (\dd{real}). The eight outlets are shown in Appendix Figure~\ref{fig:roster}. Real-logo cells are shown under nominative use; logo assets are release-unsafe and not redistributed.}
\label{fig:stim}
\end{figure}

\smallskip\noindent\textbf{Design.}
All stimuli are self-rendered news-article screenshots ($1024\times768$) from an HTML$\rightarrow$PNG pipeline, composed from two factors (Figure~\ref{fig:stim}). The \textbf{skin} is the page styling: a \emph{verified} broadsheet layout (carrying the four high-credibility brands) or a \emph{tabloid} red-top layout (carrying the four low-credibility brands), with generic mastheads in the no-brand controls. The \textbf{logo level} controls how brand identity is shown and isolates the channel under test: \dd{none} (no masthead; pure layout baseline), \dd{fictional} (a styled wordmark with a made-up name; controls for \emph{a masthead exists}), \dd{ocr\_text} (the same wordmark styling but the \emph{real brand name as text}; isolates the brand-\emph{name string} channel), and \dd{real} (the \emph{real brand logo image} on a white chip with the name hidden; isolates the brand-\emph{logo image} channel). The \dd{none} and \dd{fictional} levels are the controls that license attributing the effect to brand identity rather than to layout or to the mere presence of a masthead.

\smallskip\noindent\textbf{Brands, content, and readout.}
We use eight outlets with a fixed brand$\rightarrow$skin mapping (high: \emph{The New York Times, Reuters, The Guardian, BBC}; low: \emph{Daily Mail, Daily Mirror, The Sun, National Enquirer}) and six neutral articles on mundane, non-political topics (infrastructure, science, technology, environment, transport, public health), written in an inverted-pyramid register so topic does not drive credibility. Formally, each stimulus is a deterministic render $x=\mathcal{R}(b,a,\ell)$ of a brand $b\in\mathcal{B}$ ($|\mathcal{B}|{=}8$), an article $a\in\mathcal{A}$ ($|\mathcal{A}|{=}6$), and a logo level $\ell\in\{\textsc{none},\textsc{fict},\textsc{ocr},\textsc{real}\}$. The credibility readout is the model's last-token log-odds
\begin{equation}
\mathrm{cred}(x)\;=\;z_{\textit{Yes}}(x)-z_{\textit{No}}(x),
\label{eq:cred}
\end{equation}
where $z_{\textit{tok}}(x)$ is the answer-position logit of token \textit{tok} for \emph{Is this a credible, trustworthy news source? Answer with one word: Yes or No.} (a 1--7 self-report was degenerate and is dropped). The per-brand prior fixes the article and channel and averages over articles, $\pi(b)=\mathbb{E}_{a\in\mathcal{A}}[\mathrm{cred}(\mathcal{R}(b,a,\textsc{ocr}))]$, and the high$-$low tier gap is $\Delta=\tfrac14\sum_{b\in\mathrm{hi}}\pi(b)-\tfrac14\sum_{b\in\mathrm{lo}}\pi(b)$. The readout is a single logit contrast: its \emph{magnitude} is prompt-dependent (\S\ref{sec:robust}), while its sign, direction, and brand ordering are stable and externally validated against MBFC ratings (\S\ref{sec:phenom}). For the content analyses (\S\ref{sec:override}) we add a content-only \emph{cite} probe (\emph{Does this article cite specific named sources, records, or figures?}).

\smallskip\noindent\textbf{Released stimulus suite (contribution).}
The suite is a factorial, control-equipped benchmark for source-vs-content effects in VLMs: the main set is $8$ brands $\times$ $6$ articles across logo levels; a content-evidence set (\S\ref{sec:override}) crosses each brand with \emph{well-sourced} and \emph{red-flag} versions of the same facts (length-matched within $8\%$, differing only in epistemic markers); and a cross-lingual set translates the articles to Chinese and Spanish (brand names kept as proper nouns, question language-matched). Skins are luminance-matched and share body font size (audited), and every set is frozen by a content hash for reproducibility. The IP-clean levels (\dd{none}/\dd{fictional}/\dd{ocr\_text}) and the rendering pipeline are released; real brand logos and real screenshots are trademark/copyright \emph{release-unsafe} (used to run experiments, never redistributed) and we provide the renderer plus a brand$\rightarrow$skin specification so users supply their own assets. A datasheet is given in Appendix~\ref{app:datasheet}.

\smallskip\noindent\textbf{Analysis methods.}
We use four analyses, detailed in the appendix. \emph{Behavioral readouts} are the cred (and cite) log-odds above. \emph{Linear probing} regresses the cred log-odds from the last-token residual stream at each layer with leave-one-brand-out (LOBO) cross-validation, paired with a Hewitt--Liang control task to expose string shortcuts (Appendix~\ref{app:probe}). \emph{Activation patching} transplants the last-token residual from a donor into a recipient at each layer using matched-layout contrasts so only the masthead (or, in \S\ref{sec:override}, the content) differs. Writing $h^{(l)}(x)$ for the layer-$l$ last-token residual, patching donor $x_d$ into recipient $x_r$ gives the normalized transfer
\begin{equation}
T^{(l)}=\frac{\mathrm{cred}\big(x_r;\,h^{(l)}\!\leftarrow\!h^{(l)}(x_d)\big)-\mathrm{cred}(x_r)}{\mathrm{cred}(x_d)-\mathrm{cred}(x_r)},
\label{eq:transfer}
\end{equation}
so $T^{(l)}{=}0$ marks a causally inert layer and $T^{(l)}{=}1$ a layer that fully carries the effect (Appendix~\ref{app:patch}). \emph{Sparse autoencoders} are trained on the L20 last-token residual (top-$k$, $\times 8$ expansion); we select credibility features by correlation to the cred readout and verify them by signed feature steering, with 5-seed stability (Appendix~\ref{app:sae}). All headline numbers carry bootstrap 95\% CIs ($n{=}5000$) and permutation tests ($n{=}10000$).

\section{Mechanism: The Brand Prior, Its Locus, and Its Content Override}\label{sec:phenom}\label{sec:mech}\label{sec:override}
\smallskip\noindent\textbf{Visual source form matters, which motivates studying identity.}
Before isolating brand identity we confirm that source \emph{form} moves the readout at all and that it can dominate conflicting cues (Appendix~\ref{app:gate0}). A manipulation check shows the verified skin reads more credible than the tabloid skin with disjoint CIs; the effect is partly visual (a VLM-minus-text-LM gap, carried mainly by a social-card layout that lowers credibility by $2.2$ log-odds); and in a conflict design where the visual tier contradicts an in-image text label, an authoritative broadsheet visual overrides the conflicting label $6/6$ while a tabloid visual is pulled back by the label ($2/12$), an asymmetry ($0.83$) that recurs throughout this paper. These effects establish that the model attends to source presentation; the rest of the paper asks what the controlling cue is (brand identity), where it lives, and what it overrides.

\begin{figure}[t]
\centering
\includegraphics[width=\linewidth]{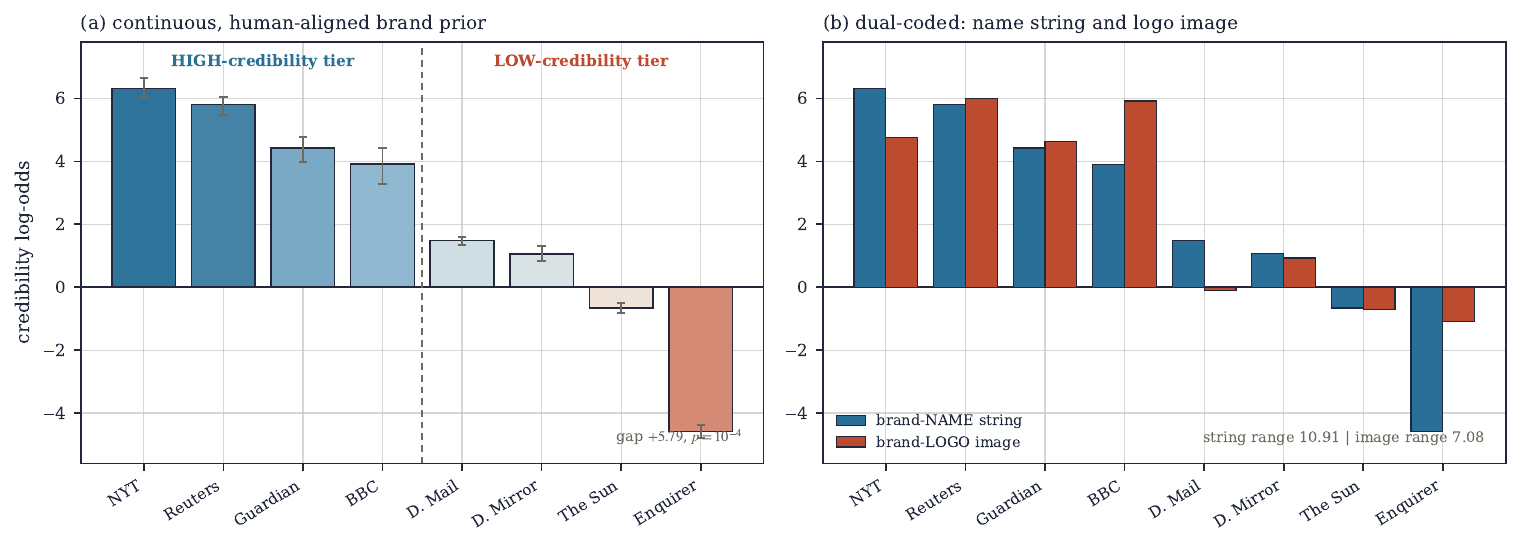}
\caption{The credibility phenomenon (Qwen2.5-VL-7B, 95\% CIs). (a) The brand prior is continuous and human-aligned: swapping only the masthead spans $10.91$ log-odds with cleanly separated high/low tiers (gap $+5.79$, $p=10^{-4}$), coloured by value on the diverging scale. (b) The prior is dual-coded: the brand-name string and the brand-logo image independently recover the ordering (string range $10.91$, image range $7.08$; the \emph{BBC} logo outscores its name, the \emph{Enquirer} name outscores its logo).}
\label{fig:phenom}
\end{figure}

\smallskip\noindent\textbf{The prior is continuous and human-aligned.}
Holding layout, content, and wordmark styling fixed and varying only the brand-name string, the credibility readout spans $10.91$ log-odds, from \emph{NYT} $+6.32$ $[6.04,6.66]$ to \emph{National Enquirer} $-4.59$ $[-4.78,-4.38]$ (Figure~\ref{fig:phenom}a, Table~\ref{tab:prior}). The ordering matches an independent professional ground truth: the prior correlates with Media Bias/Fact Check (MBFC) factual-reporting credibility at Spearman $\rho{=}0.88$ ($p{=}0.004$; Pearson $r{=}0.89$), with a wide bootstrap CI $[0.26,1.00]$ that reflects the eight-brand set and a hard permutation $p$ (Appendix~\ref{app:mbfc}). The high$-$low tier gap is $+5.79$ log-odds (permutation $p=10^{-4}$). The effect is not a tokenization artifact: common-bigram brands (\emph{The Sun} $-0.66$) and unique strings (\emph{Enquirer}) both behave consistently with reputation, and the per-brand intervals are tight across the six topics. The correlation is informatively \emph{imperfect}: the model overrates the \emph{Daily Mail} ($+1.48$, although MBFC rates it Low/Questionable) and ranks it above the \emph{Daily Mirror} (MBFC ranks the Mirror higher), and the \emph{BBC} sits behind the \emph{Guardian}. These are concrete, measurable miscalibrations (precisely the failure the framing targets, a mostly-warranted prior with local errors rather than a blanket bias) and we treat them, not the existence of the prior, as the calibration question.

\begin{table}[t]
\centering\small
\caption{Per-brand credibility log-odds with bootstrap 95\% CIs (brand-name string channel, Qwen2.5-VL-7B).}
\label{tab:prior}
\begin{tabular}{lcc@{\hskip 2em}lcc}
\toprule
Brand & cred & 95\% CI & Brand & cred & 95\% CI\\
\midrule
The New York Times & $+6.32$ & $[+6.04,+6.66]$ & Daily Mail & $+1.48$ & $[+1.33,+1.61]$\\
Reuters & $+5.80$ & $[+5.48,+6.05]$ & Daily Mirror & $+1.06$ & $[+0.83,+1.31]$\\
The Guardian & $+4.42$ & $[+3.97,+4.78]$ & The Sun & $-0.66$ & $[-0.83,-0.50]$\\
BBC & $+3.91$ & $[+3.27,+4.42]$ & National Enquirer & $-4.59$ & $[-4.78,-4.38]$\\
\bottomrule
\end{tabular}
\end{table}

\smallskip\noindent\textbf{The prior is dual-coded.}
The prior fires through two separable channels (Figure~\ref{fig:phenom}b). The brand-name string spans $10.91$ log-odds and the brand-logo image (the real logo alone, name hidden) spans $7.08$, both ordered by reputation. Route strength is brand-dependent: iconic-logo brands lean visual (the \emph{BBC} logo exceeds its name, $+5.92$ vs $+3.91$), distinctive-string brands lean textual (\emph{Enquirer} name $-4.59 \gg$ logo $-1.08$). This visual channel is the qualitative difference from text-only source-preference findings: a rendered logo, carrying no readable brand text, already recovers the credibility ordering, so the prior is keyed on \emph{learned visual identity}, not merely on a name token. Per-brand values are tabulated in Appendix~\ref{app:image}.

\smallskip\noindent\textbf{The prior scales with model size.} Across seven models and two families \citep{marafioti2025smolvlm,zhu2025internvl3,bai2025qwen25vl,li2024llavaonevision}, the prior strengthens on two axes (Table~\ref{tab:scaling}): magnitude (credibility range) grows with size, and the ordering converges to human consensus (Spearman $\rho\rightarrow1.0$). Spearman$(\text{size},\text{range})=+0.93$ (permutation $p=0.0035$); the InternVL family forms a monotone three-point ladder ($2.17\rightarrow4.62\rightarrow12.89$ at 2B/8B/14B), and at fixed size the prior is dominated by pretraining corpus rather than parameter count (LLaVA-OV-7B range $2.27$ vs Qwen-7B $10.90$). That the \emph{ordering}, not only the magnitude, sharpens with scale disfavors a pure \emph{larger models are just louder} reading.

\begin{figure}[t]
\centering
\includegraphics[width=0.95\linewidth]{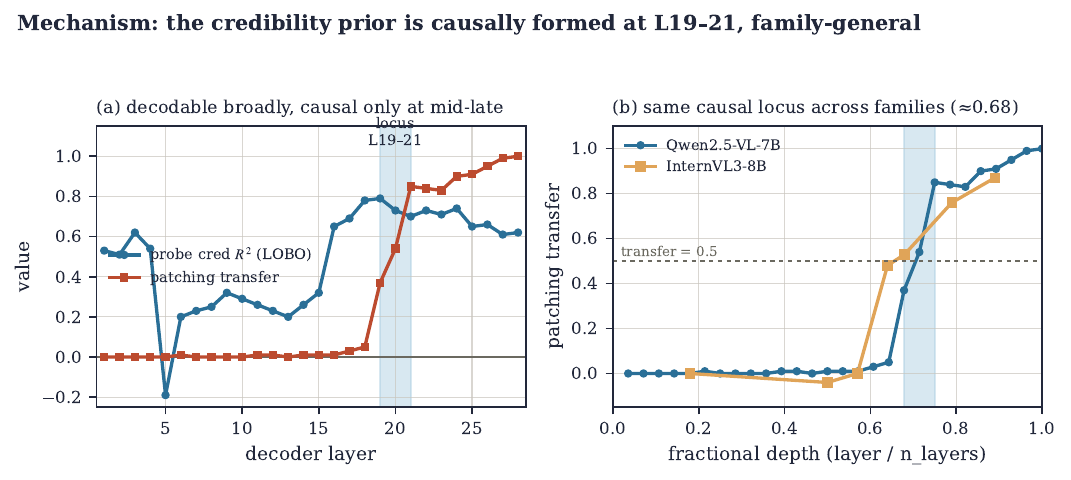}
\caption{Mechanism. (a) The cred-regression probe $R^2$ peaks at L19 while causal patching transfer onsets at L19 and saturates late: decodable broadly, causal at L19--21. (b) The same causal locus ($\approx$0.68 fractional depth) recurs in InternVL3-8B.}
\label{fig:mech}
\end{figure}

\smallskip\noindent\textbf{Where the prior lives and what it is made of.}
\smallskip\noindent\textbf{Localization.}
A naive high/low linear probe reaches perfect accuracy at every layer, including layer~1, a string shortcut exposed by a control-task selectivity test \citep{hewitt2019control}, where probes fit random-but-fixed brand labels at mean accuracy $0.64$ (selectivity $\approx+0.36$). We therefore localize with a \emph{continuous} cred-regression probe (leave-one-brand-out ridge) and with causal activation patching. The cred-regression $R^2$ rises sharply through the mid layers and peaks at L19 ($R^2=0.79$); matched-layout last-token activation patching (donor \emph{Daily Mail} $\rightarrow$ recipient \emph{Enquirer}, six matched articles) shows transfer near zero through the first $\sim$60\% of layers, a sharp onset at L19 ($0.37$), a sign-flip by L21, and saturation in the late layers (Figure~\ref{fig:mech}a). Credibility is thus \emph{decodable broadly but causally formed only at L19--21} (the full probe$\rightarrow$patch$\rightarrow$steer procedure is Algorithm~\ref{alg:tda}).

\smallskip\noindent\textbf{Cross-family locus.}
Porting the patching analysis to InternVL3-8B (a different family, same depth) places the causal locus at fractional depth $0.68$ (essentially identical to Qwen's L19--21) with the same sharp mid-late onset (Figure~\ref{fig:mech}b). Two independently trained, architecturally distinct families place the credibility causal locus at the same relative depth: the mechanism, not merely the phenomenon, generalizes.

\begin{figure}[t]
\centering
\includegraphics[width=0.95\linewidth]{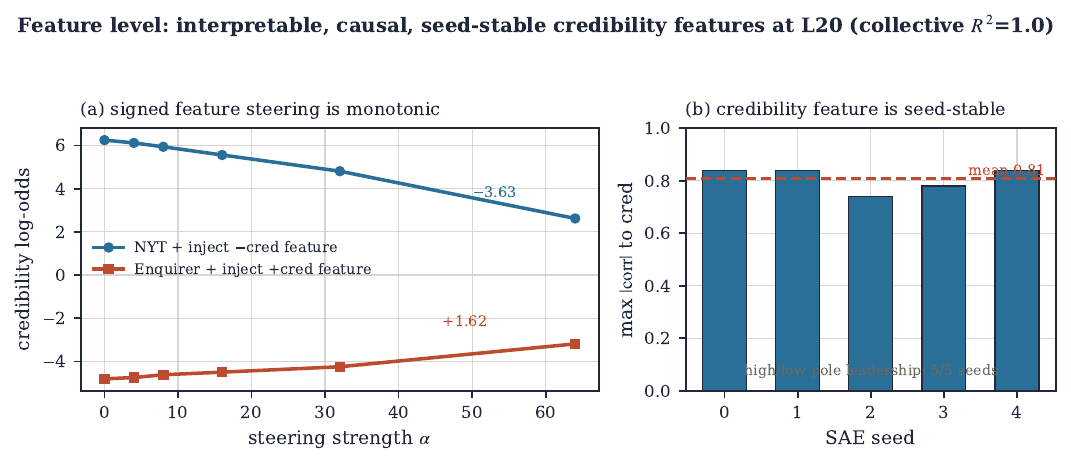}
\caption{Feature-level mechanism at L20. (a) Signed SAE-feature steering moves the credibility readout monotonically in both directions. (b) The credibility feature is seed-stable (mean $\max|\mathrm{corr}|=0.81$; 5/5 pole leadership).}
\label{fig:sae}
\end{figure}

\smallskip\noindent\textbf{Feature level.}
A single-layer top-$k$ sparse autoencoder at L20 ($\times8$ expansion, $k{=}32$) linearly encodes the credibility direction: the cred readout is recoverable from the L20 features (collective $R^2{=}1.0$). We treat this recoverability as a sanity check (with many features such a fit is expected) and rest the feature-level claim on the causal steering and 5-seed stability below, not on $R^2$. It contains clean, monosemantic credibility features: a $+$cred feature fires on \emph{NYT}/\emph{Reuters}/\emph{BBC}, a $-$cred feature on \emph{Enquirer}/\emph{Sun}/\emph{Mirror}. Signed feature steering moves the readout monotonically, injecting the $-$cred feature into \emph{NYT} drops it $-3.63$ log-odds; injecting the $+$cred feature into \emph{Enquirer} raises it $+1.62$ (Figure~\ref{fig:sae}a), establishing feature-level causality, not mere correlation. The structure is seed-stable: across five SAEs the maximum credibility correlation averages $0.81$ and the poles are led by high/low-authority brands in 5/5 seeds (Figure~\ref{fig:sae}b); feature \emph{indices} vary by seed but the high$\rightarrow$positive, low$\rightarrow$negative organization does not. A recurring asymmetry (pulling a high brand down is easier than pushing a low brand up, and the $-$cred pole is the more stable of the two) indicates that low credibility is written more robustly into the representation, consistent with the override results below.

\smallskip\noindent\textbf{The prior overrides content.}
We cross brand tier with content-evidence quality (\emph{well-sourced} vs \emph{red-flag}: identical facts and named source, differing only in epistemic markers, named attributions and public records vs anonymous \emph{sources say} and sensational framing).

\begin{figure}[t]
\centering
\includegraphics[width=0.97\linewidth]{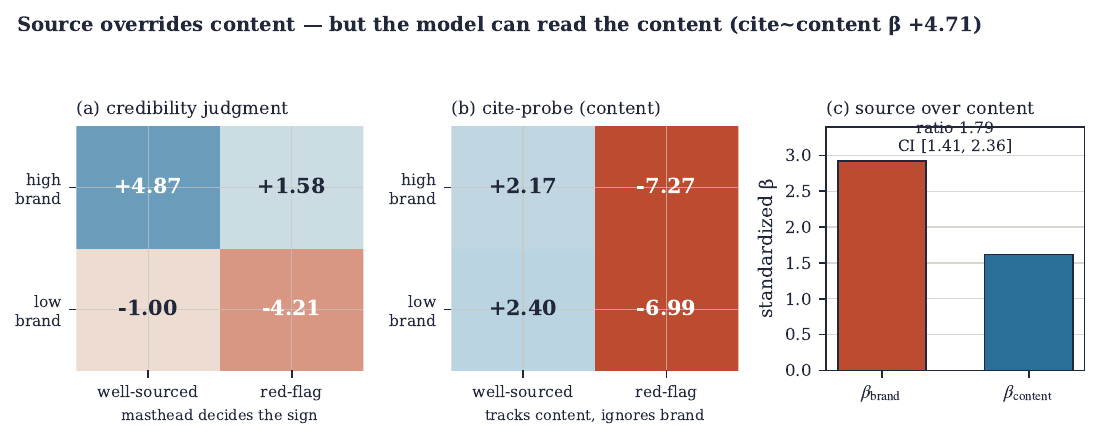}
\caption{Source overrides content. (a) Credibility splits by brand row, the masthead decides the sign in conflict cells. (b) The cite-probe splits by content column, the model reads content and ignores brand. (c) Standardized regression: $\beta_{\text{brand}}/\beta_{\text{content}}=1.79$, CI excludes 1.}
\label{fig:override}
\end{figure}

\smallskip\noindent\textbf{The masthead decides the sign.}
In the $2\times2$ credibility table (Figure~\ref{fig:override}a) the conflict cells are decided by the source: a high brand with red-flag content is still judged credible ($+1.58$), a low brand with well-sourced content is still judged not credible ($-1.00$). A standardized regression
\begin{equation}
\mathrm{cred}=\beta_0+\beta_{\text{brand}}\,\tau(b)+\beta_{\text{content}}\,e(c)+\varepsilon,
\label{eq:override}
\end{equation}
with brand tier $\tau(b)$ and evidence quality $e(c)$ both $z$-scored, gives $\beta_{\text{brand}}=2.92$ ($p=10^{-4}$) vs $\beta_{\text{content}}=1.62$ ($p=2\times10^{-4}$); the magnitude ratio is $1.79$ with 95\% CI $[1.41,2.36]$ (excludes~1). Source outweighs content $\sim$1.8$\times$.

\smallskip\noindent\textbf{This is deference, not a reading deficit.}
On the content-only \emph{cite} probe the model tracks content and ignores brand ($\beta_{\text{content}}=+4.71$, $\beta_{\text{brand}}=-0.13$; Figure~\ref{fig:override}b): it \emph{can} read sourcing quality but declines to use it for credibility, deferring to the prior. Content shifts every brand additively by a similar amount ($+3.29$ high vs $+3.21$ low), so conflict cells are decided by the prior only because the brand baseline gap exceeds the content shift, not because low brands \emph{resist} good content (the \emph{Enquirer} is the exception, staying negative even when well-sourced).

\smallskip\noindent\textbf{Mechanism of the override.}
Two matched patching contrasts, a brand axis (content fixed) and a content axis (brand fixed), transfer with \emph{equal} efficiency at L19--21 ($0.76$ vs $0.75$; difference CI $[-0.03,+0.05]$, Appendix~\ref{app:p4}): brand and content converge on one shared mechanism, not competing routes. The override is thus a \emph{signal-magnitude} effect, the brand injects a larger representational difference (gap $10.4$ vs $3.6$ log-odds) into a shared readout, which is why good content cannot rescue a low-credibility brand. SAE localization of the override was inconclusive (credibility is distributed across L20 features), which is why we answer this with patching.

\smallskip\noindent\textbf{The prior is causally controllable.}\label{sec:intervention}
The same $-$credibility SAE direction that steers the readout (\S\ref{sec:mech}) can intervene on the override. Steered into high-brand stimuli only (Figure~\ref{fig:tradeoff}a), it cuts the conflict-cell override gap by $41\%$ ($5.79\rightarrow3.44$) while retaining $75\%$ of content sensitivity and flips the good-brand$+$red-flag cell from $+1.58$ to $-0.76$. It is \emph{selective}, the untouched low-brand control cell stays at $-4.21$; pushed further it cleanly neutralizes the high$-$low gap to non-significance ($+5.79\rightarrow+0.23$ at $\alpha{=}96$, Figure~\ref{fig:tradeoff}b) without inversion. The direction is not overfit to our eight brands: applied unchanged to seven held-out outlets (e.g.\ NPR, Al Jazeera, InfoWars) it moves their credibility at $61\%$ of the in-sample slope (Appendix~\ref{app:heldout}), so it is a general brand-credibility axis. We frame it as an interpretable controllability handle, not the strongest suppressor: a one-sentence \emph{judge by the evidence, ignore the outlet} prompt reduces the override further ($-89\%$ vs $-41\%$, Appendix~\ref{app:promptmit}), but that prompt is black-box and on/off, whereas the steering direction is interpretable, continuously dialable, selective, and independent of instruction-following. The comparison is itself informative, the prompt \emph{raises} content sensitivity rather than adding a capability, corroborating that the override is a deployment-default gap and not a missing ability. Because the single direction captures the global brand axis, it turns the prior off wholesale rather than re-calibrating individual brands; this measure$\rightarrow$localize$\rightarrow$intervene chain is not available to behavioral-only source-preference studies.

\section{CueTrust: A Cross-Model Source-Override Benchmark}\label{sec:cuetrust}
The brand override generalizes cleanly along one dimension and stops sharply along another. We formalize it as the \emph{Source-Override Index}, from the standardized regression (Eq.~\ref{eq:override}) fit per cue axis,
\begin{equation}
\mathrm{SOI}=|\beta_{\text{cue}}|/|\beta_{\text{content}}|,
\label{eq:soi}
\end{equation}
a within-model ratio ($\mathrm{SOI}{>}1$ means the cue moves credibility more than the article's evidence) that is comparable across models despite differing logit scales. The five cues span three encodings of outlet identity (name, logo, domain) and three non-outlet negative controls (author, in-text authority, layout), each a $2{\times}2$ cue$\times$content design; sweeping them over seven models (Algorithm~\ref{alg:soi}) gives the CueTrust matrix (Table~\ref{tab:cuetrust}). Construction, design rationale, and statistics are in Appendix~\ref{app:cuetrust}.

\begin{table}[t]
\centering\small
\setlength{\tabcolsep}{6pt}
\caption{\textbf{CueTrust.} Source-Override Index (SOI $=|\beta_{\text{cue}}|/|\beta_{\text{content}}|$) for five source cues across seven VLMs. Bold $=$ SOI $>1$ (cue overrides content; brand and domain effects $p\le5\times10^{-4}$); author and authority are null across all models (negative controls).}
\label{tab:cuetrust}
\begin{tabular}{lccccc}
\toprule
Model & brand & domain & author & format & authority\\
\midrule
Qwen2.5-VL-3B & \textbf{3.44} & \textbf{1.93} & $0.05$ & $0.37$ & $0.16$\\
Qwen2.5-VL-7B & \textbf{1.79} & \textbf{1.49} & $0.06$ & $0.28$ & $0.10$\\
InternVL3-2B & \textbf{1.27} & $0.24$ & $0.09$ & \textbf{1.12} & $0.03$\\
InternVL3-8B & \textbf{1.64} & \textbf{1.33} & $0.28$ & $0.78$ & $0.07$\\
InternVL3-14B & \textbf{1.34} & \textbf{3.13} & $0.55$ & $0.21$ & $0.07$\\
LLaVA-OV-7B & $0.92$ & $0.64$ & $0.05$ & $0.94$ & $0.09$\\
SmolVLM-2B & $0.44$ & $0.06$ & $0.36$ & $0.24$ & $0.12$\\
\bottomrule
\end{tabular}
\end{table}

\smallskip\noindent\textbf{The override is outlet-identity-specific and encoding-invariant.} Three encodings of the publishing outlet's identity trigger it, the masthead \emph{name} (SOI $1.79$), the \emph{logo} image (an independent channel, \S\ref{sec:phenom}), and the bare \emph{domain}/URL with no logo or masthead (SOI $1.45$, $p{=}2\times10^{-4}$); three non-outlet cues do not, an in-text \emph{authority} attribution (SOI $0.08$), the page \emph{format} including a strong social-media UI (SOI $0.13$--$0.28$), and a named \emph{author} byline (SOI $0.07$). The prior is keyed on \emph{which outlet} published the article and is invariant to how that identity is presented, but does not extend to author identity, in-text attribution, or layout, which are absorbed as content. Authority, format, and author are therefore clean negative controls: the override is not a reaction to any authoritative-looking cue, only to outlet identity, and domain is a third encoding-invariant channel alongside the name and logo of \S\ref{sec:phenom}.

\smallskip\noindent\textbf{Vulnerability profiles are model- and scale-dependent.} Brand overrides most often but varies $\sim$8$\times$ (Qwen-3B $3.44$ to SmolVLM $0.44$, which overrides on nothing); domain reliance grows monotonically with scale inside InternVL ($0.24\rightarrow1.33\rightarrow3.13$), while format is a small-model vulnerability (InternVL-2B $1.12$, LLaVA $0.94$) that vanishes at scale, so the dominant cue shifts from layout to source identity as models grow. CueTrust is thus a deployable diagnostic reading a candidate VLM's source-cue vulnerability profile, a dimension detection-accuracy benchmarks do not measure.

\section{Discussion}\label{sec:discuss}
\smallskip\noindent\textbf{Robustness.}\label{sec:robust} The brand prior survives five stress tests (Appendix~\ref{app:prompt}, Figure~\ref{fig:robust}): it is present under all five credibility prompts (a polarity-flip and a forced-choice token set included, ruling out a yes-bias and a Yes/No-token artifact; magnitude is prompt-dependent, existence and direction are not), in Chinese and Spanish ($\rho\geq0.98$; keyed on identity, not English text), at the \emph{same} causal locus in a second model family (\S\ref{sec:mech}), and on real first-fold screenshots of seven live news sites (range $9.25$, $\rho=0.96$), which closes the synthetic-only critique. Every headline result carries bootstrap CIs and permutation tests (brand gap $p=10^{-4}$, override ratio CI excludes~1, scaling $p=0.0035$).
\smallskip\noindent\textbf{Limitations.} Scaling uses seven models (Spearman CI lower bound $0.32$; permutation $p$ robust). We localize the override with patching, not a transcoder attribution graph, because SAE feature-localization was inconclusive (credibility is \emph{distributed} across L20 features) \citep{dunefsky2024transcoders,lindsey2025biology}. The main set is synthetic and single-domain; cross-lingual and real-screenshot results mitigate but do not eliminate this, and Layer~B is 7/8 brands on one model.
\smallskip\noindent\textbf{Implication.} Deployed VLMs reading news imagery may defer to source identity over the evidence present; because the model \emph{can} read that evidence, the failure is one of deference, not perception, relevant wherever VLMs sit in fact-checking or moderation pipelines.
\smallskip\noindent\textbf{Threat model.} The override is exploitable both ways: an adversary can \emph{launder} weak content under a trusted masthead (high$+$red-flag stays credible, $+1.58$) or \emph{suppress} strong content under a distrusted one ($-1.00$). Because the cue is a rendered identity, not a verifiable provenance signal, a pipeline treating the VLM's judgment as evidence-grounded inherits a source channel it did not intend to trust.

\smallskip\noindent\textbf{Conclusion.} VLMs trust the masthead over the evidence, the cue is carried visually as well as textually, and we have shown where in the network that judgment lives and why source wins.

\section*{Reproducibility Statement}
All experimental details needed to reproduce our results are provided. The stimulus construction, brand$\rightarrow$skin mapping, article topics, and frozen content hashes are in Appendix~\ref{app:datasheet}; the full per-layer probe diagnostics and activation-patching curves (all 28 layers, both model families) in Appendices~\ref{app:probe}--\ref{app:patch}; the sparse-autoencoder configuration, $\alpha$-sweep, and five-seed feature tables in Appendix~\ref{app:sae}; the source-override design and standardized regression in Appendix~\ref{app:p4}; the external MBFC ground-truth mapping in Appendix~\ref{app:mbfc}; and the models, quantization, and statistical protocol (bootstrap $n{=}5000$, permutation $n{=}10000$) at the end of the appendix. The IP-clean stimulus levels, the renderer, the brand$\rightarrow$skin specification, and all figure- and analysis-generation code are released as supplementary material, so every synthetic result can be regenerated end-to-end. Real brand logos and real news screenshots are trademarked and release-unsafe and are therefore not redistributed; the renderer accepts user-supplied logo assets to reproduce those cells. MBFC ratings correspond to the version accessed June 2026 and should be re-pinned to a dated snapshot.

\section*{Ethics Statement}
This work studies a reliability failure, not an attack. It involves no human subjects and no private data. The released stimuli are synthetic; we deliberately withhold real brand logos and screenshots to respect trademark and copyright. Our threat model (\S\ref{sec:discuss}) describes how a source-identity channel could be exploited to launder or suppress content; we disclose it because the same mechanism and locus are what make detection and mitigation possible, and because deployed VLMs already exhibit the behavior. Credibility here is operationalized against Media Bias/Fact Check, a third-party rating that is itself contested and Anglophone-centric; we treat the model--MBFC correlation as evidence of alignment with one professional standard, not as ground truth about any outlet, and we report the model's miscalibrations explicitly. The eight-outlet, English-first scope limits the generality of any claim about which sources are trustworthy.

\section*{Use of Large Language Models}
Large language models were used as a general-purpose assistant for drafting and editing the manuscript and for writing figure- and analysis-generation code. All experimental design, results, numerical findings, and scientific claims are the authors' own and were verified by the authors; external references and third-party ratings were checked against their original sources.

\bibliography{references}
\bibliographystyle{iclr2026_conference}

\appendix
\section{Dataset, Stimuli, and Reproducibility}

\smallskip\noindent\textbf{Dataset datasheet.}\label{app:datasheet}
\textbf{Motivation.} A controlled benchmark for separating source identity from layout and from content evidence in VLM credibility judgments. \textbf{Composition.} Eight outlets (high: NYT, Reuters, Guardian, BBC; low: Daily Mail, Daily Mirror, The Sun, National Enquirer) on a fixed brand$\rightarrow$skin mapping; six neutral articles (\dd{a01\_reservoir}, \dd{a02\_telescope}, \dd{a03\_battery}, \dd{a04\_coral}, \dd{a05\_rail}, \dd{a06\_vaccine\_storage}) in inverted-pyramid register on non-political topics. \emph{Main set}: $8{\times}6$ across logo levels. \emph{Content-evidence set}: each article additionally has \emph{well-sourced} and \emph{red-flag} versions, same facts and named source, length-matched within $8\%$, differing only in epistemic markers (named attributions $+$ public-register records $+$ measured hedging vs.\ anonymous \emph{sources say} $+$ no records $+$ sensational framing). \emph{Cross-lingual set}: the six articles translated to Chinese and Spanish, brand names kept as proper nouns, question language-matched. \textbf{Collection process.} Synthetic; HTML$\rightarrow$PNG rendering at $1024{\times}768$ via templated skins; skins luminance-matched and sharing body font size, audited and frozen by content hash. \textbf{Distribution.} The \dd{none}/\dd{fictional}/\dd{ocr\_text} levels, the content-evidence and cross-lingual sets, the renderer, and the brand$\rightarrow$skin specification are released (synthetic content under an open license). Real brand logos (\dd{real} level) and real news screenshots are trademark/copyright \emph{release-unsafe} and are not redistributed; the renderer lets users supply their own logo assets. \textbf{Limitations.} Single news domain; eight outlets; the real-screenshot validation covers 7/8 brands on one model.

\begin{table}[h]\centering\small
\caption{Frozen stimulus sets (content hashes for reproducibility).}
\begin{tabular}{llll}
\toprule
hash & $n$ & set & use\\
\midrule
\dd{d803eed8} & 144 & behavioral & foundational battery\\
\dd{272495f1} & 180 & logo ablation & channel isolation\\
\dd{06d67e20} & 288 & 8-brand \dd{ocr\_text} & brand prior\\
\dd{52930fe8} & 396 & 8 real logos & current frozen main set\\
\dd{e32b96e4} & 96 & brand$\times$content & source override\\
\dd{xling\_zh/es} & 48/48 & cross-lingual & robustness\\
\dd{stimuli\_realB} & 21 & real screenshots & ecological validity, release-unsafe\\
\bottomrule
\end{tabular}
\end{table}

\begin{figure}[t]
\centering
\includegraphics[width=0.95\linewidth]{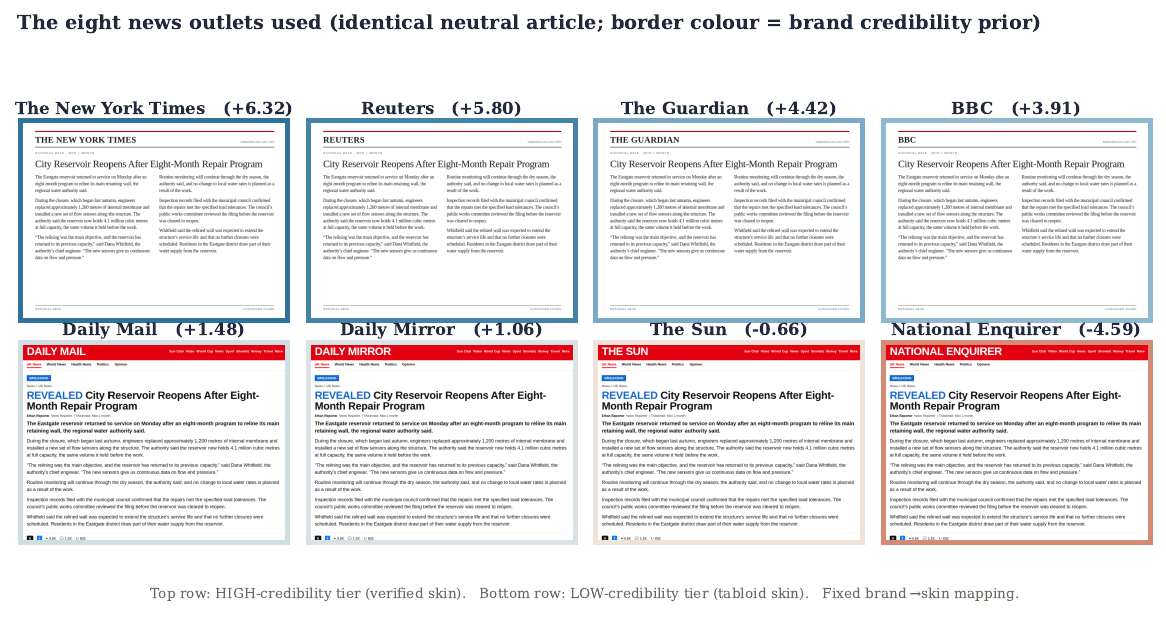}
\caption{The eight news outlets, one example article each (identical neutral content). Top row: high-credibility tier on the verified skin; bottom row: low-credibility tier on the tabloid skin. Border colour encodes the brand credibility prior.}
\label{fig:roster}
\end{figure}

\section{Foundational Behavioral Battery}\label{app:gate0}
\begin{figure}[h]\centering\includegraphics[width=0.86\linewidth]{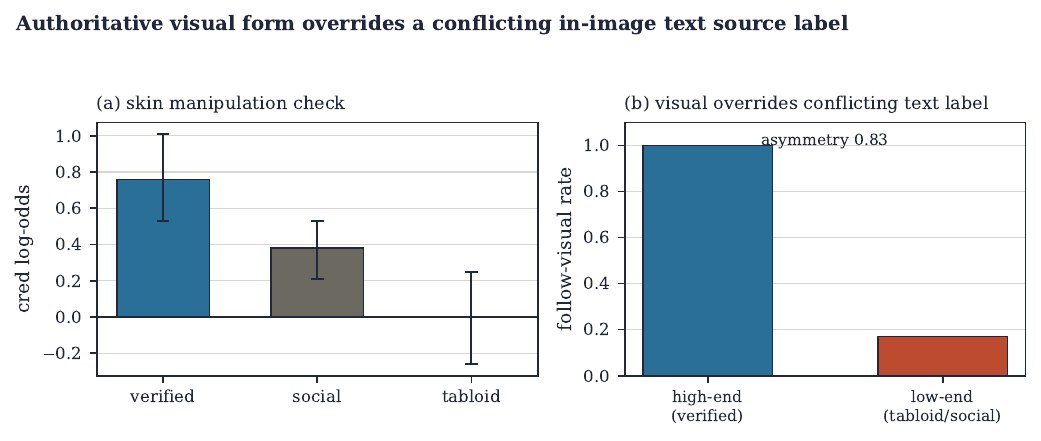}
\caption{Foundational behavioral results.}\label{fig:appgate}\end{figure}

\begin{table}[h]\centering\small
\caption{Foundational behavioral battery. (left) Skin manipulation check, neutral content, generic masthead. (right) Visual$\leftrightarrow$text-label conflict: rate of following the visual tier.}
\begin{tabular}{lcc@{\hskip 3em}lc}
\toprule
skin & cred & 95\% CI & conflict condition & follow-visual\\
\midrule
verified & $+0.76$ & $[+0.53,+1.01]$ & high-end (verified visual) & $6/6=1.00$\\
social & $+0.38$ & $[+0.21,+0.53]$ & low-end (tabloid/social) & $2/12=0.17$\\
tabloid & $0.00$ & $[-0.26,+0.25]$ & asymmetry & $0.83$\\
\bottomrule
\end{tabular}
\end{table}
The VLM$-$(text-LM) delta is $-0.84$ $[-1.31,-0.39]$ overall, carried almost entirely by the social-card layout ($-2.22$ $[-2.46,-2.03]$); broadsheet and tabloid visuals are $\approx$ their text equivalents. See Figure~\ref{fig:appgate}.

\section{Logo Ablation: Channel Isolation}\label{app:ablation}
\begin{table}[h]\centering\small
\caption{Cred log-odds by skin $\times$ logo level (left), and the two-channel decomposition for the exemplar brands (right).}
\begin{tabular}{lcccc@{\hskip 2em}lcccc}
\toprule
skin & none & fict. & \dd{ocr} & real & decomp. & generic & name & image & total\\
\midrule
verified & $+0.41$ & $+0.76$ & $+3.91$ & $+5.82$ & verified (BBC) & $+0.35$ & $+3.15$ & $+1.91$ & $+5.41$\\
tabloid & $-0.58$ & $0.00$ & $-0.66$ & $-0.89$ & tabloid (Sun) & $+0.58$ & $-0.66$ & $-0.23$ & $-0.31$\\
social & $-0.05$ & $+0.38$ & ( & ) & & & & &\\
\bottomrule
\end{tabular}
\end{table}
Layout alone is weak ($+0.41$); the brand signal dominates and the name-string and logo-image channels are separable and additive. This is the pivot from a \emph{visual source trust} framing to brand-as-memory.

\section{Probe Localization in Full}\label{app:probe}
\begin{figure}[h]\centering\includegraphics[width=0.9\linewidth]{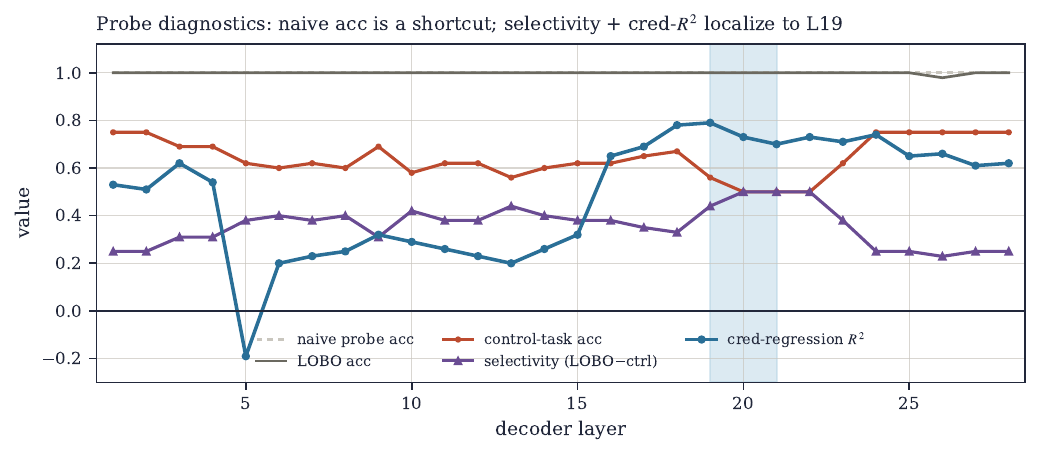}
\caption{Per-layer probe diagnostics. Naive and LOBO accuracy are saturated (shortcut); control-task accuracy, selectivity, and cred-regression $R^2$ localize the usable credibility signal to L16--19. Control-task selectivity follows \citet{hewitt2019control}.}\label{fig:appprobe}\end{figure}

A naive high/low probe reaches $1.0$ at every layer (a string shortcut). The control-task accuracy (probes fit random-but-fixed brand labels) is high (mean $0.64$), so naive accuracy is not valid localization. The continuous cred-regression $R^2$ (LOBO ridge) peaks at L19 ($0.79$). Figure~\ref{fig:appprobe} shows all five diagnostics across 28 layers.

\section{Activation Patching in Full}\label{app:patch}
Matched-layout last-token patching (donor \emph{Daily Mail} $\rightarrow$ recipient \emph{Enquirer}), six articles. Transfer is $\approx 0$ through L1--L17, onsets at L19, and saturates late; credibility flips sign by L21.
\begin{table}[h]\centering\small
\caption{(left) Qwen2.5-VL-7B patching transfer by layer (L1--16 omitted; all $\le 0.01$). (right) InternVL3-8B by fractional depth: same locus at $\approx 0.68$. Patching protocol follows \citet{zhang2024patching}.}
\begin{tabular}{lccccccc@{\hskip 2.5em}lc}
\toprule
Qwen L & 17 & 18 & 19 & 20 & 21 & 24 & 28 & IVL frac & transfer\\
\midrule
transfer & $0.03$ & $0.05$ & $0.37$ & $0.54$ & $0.85$ & $0.90$ & $1.00$ & $0.57$ & $-0.00$\\
 & & & & & & & & $0.64$ & $+0.48$\\
 & & & & & & & & $0.68$ & $+0.53$\\
 & & & & & & & & $0.79$ & $+0.76$\\
 & & & & & & & & $0.89$ & $+0.87$\\
\bottomrule
\end{tabular}
\end{table}

\section{Sparse Autoencoder in Full}\label{app:sae}
SAE config: L20 last-token residual, hidden $3584$, $\times 8$ expansion ($28672$ features), top-$k$ $k{=}32$, final FVU $0.025$. Collective cred $R^2{=}1.0$. Top features: $+$cred \dd{1299} ($+0.745$, fires NYT/Reuters/BBC); $-$cred \dd{1863} ($-0.844$, fires Enquirer/Sun/Mirror).
\begin{table}[h]\centering\small
\setlength{\tabcolsep}{4pt}
\caption{(left) Signed feature steering ($\alpha$-sweep) at L20. (right) 5-seed stability: a strong credibility feature exists in every seed and the poles are led by high/low-authority brands $5/5$. Feature steering follows \citet{arad2025saes}.}
\begin{tabular}{lcccccc@{\hskip 1.2em}lccc}
\toprule
$\alpha$ & 0 & 4 & 8 & 16 & 32 & 64 & seed & $+$corr & $-$corr & $\max|\cdot|$\\
\midrule
NYT$+(-$cred$)$ & $6.25$ & $6.12$ & $5.94$ & $5.56$ & $4.81$ & $2.62$ & 0 & $+0.74$ & $-0.84$ & $0.84$\\
Enq$+(+$cred$)$ & $-4.81$ & $-4.75$ & $-4.62$ & $-4.50$ & $-4.25$ & $-3.19$ & 1 & $+0.80$ & $-0.84$ & $0.84$\\
 & & & & & & & 2 & $+0.66$ & $-0.74$ & $0.74$\\
 & & & & & & & 3 & $+0.78$ & $-0.74$ & $0.78$\\
 & & & & & & & 4 & $+0.80$ & $-0.84$ & $0.84$\\
\bottomrule
\end{tabular}
\end{table}
\begin{figure}[h]
\centering
\includegraphics[width=0.5\linewidth]{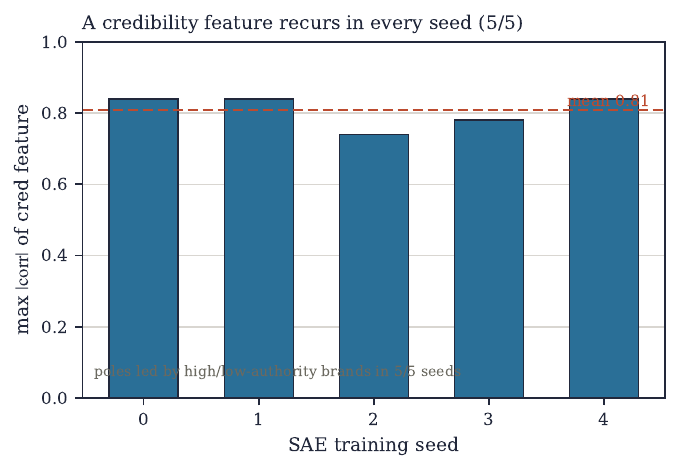}
\caption{SAE 5-seed stability: a strong credibility feature (max $|\mathrm{corr}|$ to the cred readout) recurs in every seed, with poles led by high/low-authority brands in $5/5$.}
\label{fig:saeseeds}
\end{figure}

Mean $\max|\mathrm{corr}|=0.81$. Pulling NYT down ($-3.63$) is easier than pushing Enquirer up ($+1.62$): low credibility is encoded more robustly.

\section{Source Override in Full}\label{app:p4}
\begin{figure}[h]\centering\includegraphics[width=0.74\linewidth]{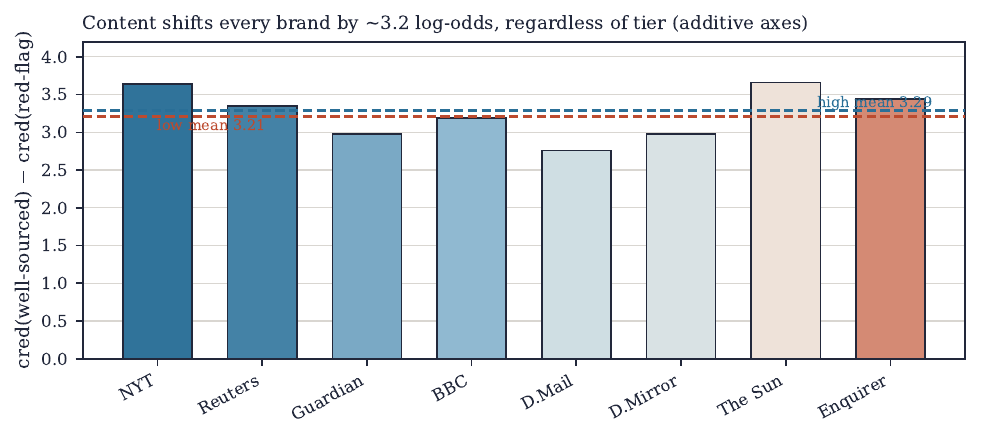}
\caption{Per-brand content sensitivity: content shifts every brand by $\approx 3.2$ log-odds regardless of tier.}\label{fig:appsens}\end{figure}

\begin{table}[h]\centering\small
\caption{Source-override $2{\times}2$ tables. Credibility splits by brand row; the cite-probe splits by content column.}
\begin{tabular}{lcc@{\hskip 3em}lcc}
\toprule
cred & well-sourced & red-flag & cite & well-sourced & red-flag\\
\midrule
high brand & $+4.87$ & $+1.58$ & high brand & $+2.17$ & $-7.27$\\
low brand & $-1.00$ & $-4.21$ & low brand & $+2.40$ & $-6.99$\\
\bottomrule
\end{tabular}
\end{table}
Standardized regression: $\beta_{\text{brand}}{=}2.92$ ($p{=}10^{-4}$), $\beta_{\text{content}}{=}1.62$ ($p{=}2{\times}10^{-4}$), ratio $1.79$ CI $[1.41,2.36]$; cite$\sim$content $\beta{=}{+}4.71$, cite$\sim$brand $\beta{=}{-}0.13$. Per-brand content sensitivity (Figure~\ref{fig:appsens}) averages $+3.29$ (high) vs $+3.21$ (low), additive across tiers. Shared-pathway per-layer transfer (brand vs content): L18 $0.46/0.46$, L19 $0.55/0.54$, L20 $0.86/0.85$, L21 $0.86/0.86$, L24 $0.92/0.93$, L26 $1.00/1.02$; difference CI at L19--21 $[-0.03,+0.05]$.

\section{Cross-lingual, Real-World, and Prompt Robustness}\label{app:prompt}
\begin{figure}[h]
\centering
\includegraphics[width=0.9\linewidth]{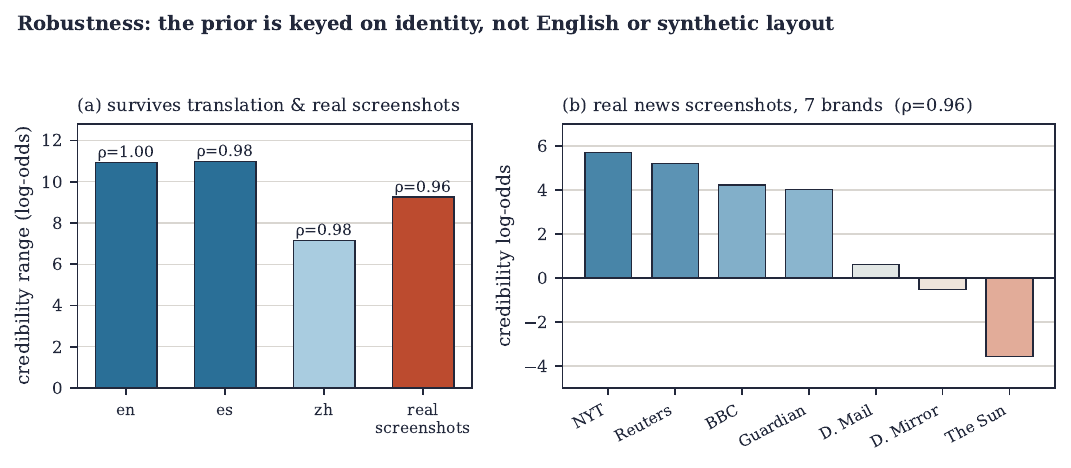}
\caption{Robustness. (a) The prior survives translation (Chinese, Spanish) and real news screenshots ($\rho\geq0.96$). (b) Per-brand credibility on real screenshots reproduces the synthetic ordering.}
\label{fig:robust}
\end{figure}
\begin{table}[h]\centering\small
\caption{(left) Cross-lingual. (middle) Real-world screenshots, per brand. (right) Prompt robustness.}
\begin{tabular}{lccc@{\hskip 1.5em}lc@{\hskip 1.5em}lcc}
\toprule
lang & range & hi$-$lo & $\rho$ & real-site brand & cred & prompt & range & $\rho$\\
\midrule
en & $10.92$ & $5.79$ & $1.00$ & NYT & $+5.69$ & v1 credible & $10.92$ & $1.00$\\
es & $10.99$ & $5.86$ & $0.98$ & Reuters & $+5.21$ & v2 reliable & $4.48$ & $0.88$\\
zh & $7.15$ & $3.22$ & $0.98$ & BBC & $+4.23$ & v3 trust & $3.30$ & $0.48$\\
 & & & & Guardian & $+4.02$ & v4 flip & $3.47$ & $0.52$\\
 & & & & Daily Mail & $+0.62$ & v5 factual & $4.58$ & $0.88$\\
 & & & & Daily Mirror & $-0.52$ & & &\\
 & & & & The Sun & $-3.56$ & & &\\
\bottomrule
\end{tabular}
\end{table}
\begin{figure}[h]
\centering
\includegraphics[width=0.95\linewidth]{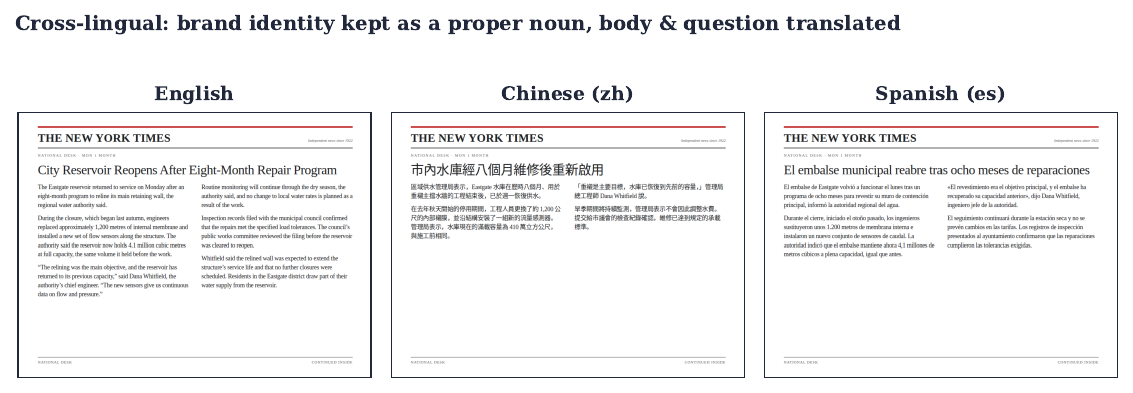}
\caption{Cross-lingual example (NYT): brand identity kept as a proper noun while body and question are translated to Chinese and Spanish.}
\label{fig:xling}
\end{figure}

\begin{figure}[h]
\centering
\includegraphics[width=0.98\linewidth]{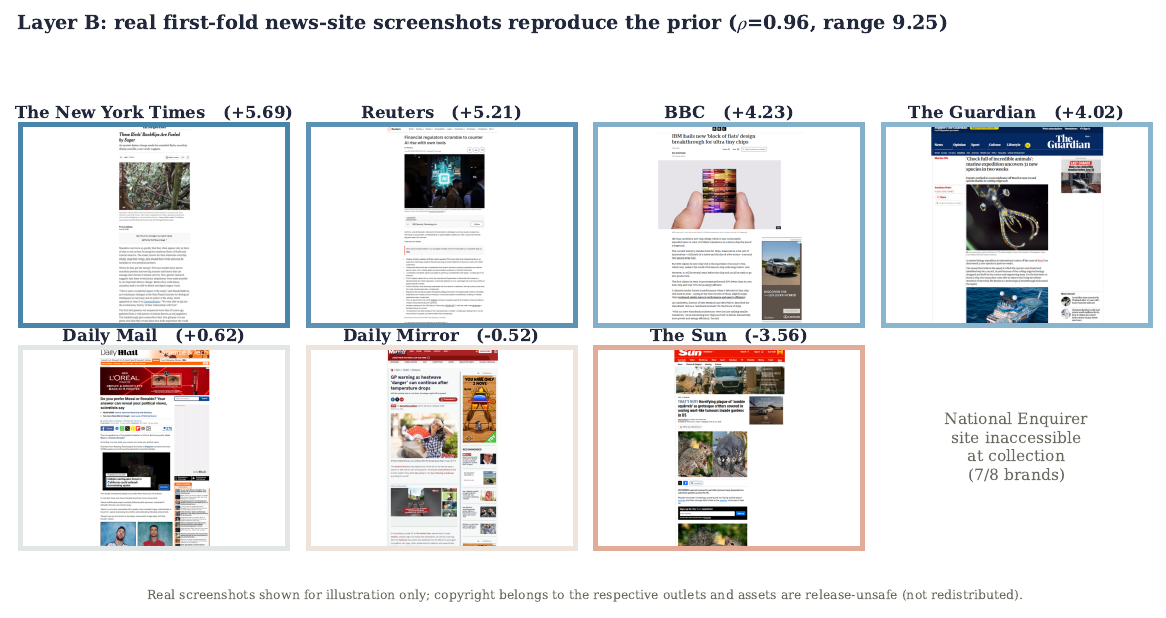}
\caption{Real-world screenshots. First-fold news-site screenshots reproduce the synthetic ordering ($\rho{=}0.96$, range $9.25$). Border colour encodes the measured credibility. Shown for illustration only; copyright belongs to the respective outlets and assets are release-unsafe (not redistributed).}
\label{fig:realB}
\end{figure}

\begin{figure}[h]
\centering
\includegraphics[width=0.55\linewidth]{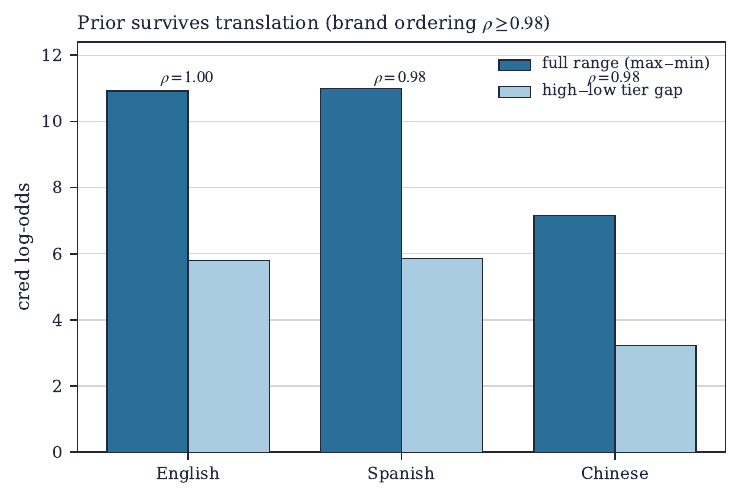}
\caption{Cross-lingual robustness: the credibility range and high$-$low tier gap in English, Spanish, and Chinese, with per-language brand-ordering agreement $\rho\geq0.98$.}
\label{fig:xlingbar}
\end{figure}

On real screenshots, range $9.25$, $\rho{=}0.96$ (Enquirer site inaccessible during collection: 7/8 brands). Prompt: mean inter-variant rank agreement $0.77$; existence and direction robust, magnitude prompt-dependent; the polarity-flipped v4 keeps NYT most credible after sign-correction (rules out a yes-bias) and v5 uses different answer tokens (rules out a Yes/No-token artifact).

\section{Brand-name vs Brand-logo Channel, per Brand}\label{app:image}
String range $10.91$, image range $7.08$.
\begin{table}[h]\centering\small
\begin{tabular}{lccc@{\hskip 2em}lccc}
\toprule
Brand & string & image & $\Delta$ & Brand & string & image & $\Delta$\\
\midrule
Reuters & $+5.80$ & $+6.00$ & $+0.20$ & Daily Mirror & $+1.06$ & $+0.94$ & $-0.12$\\
BBC & $+3.91$ & $+5.92$ & $+2.01$ & Daily Mail & $+1.48$ & $-0.12$ & $-1.59$\\
The Guardian & $+4.42$ & $+4.63$ & $+0.22$ & The Sun & $-0.66$ & $-0.67$ & $-0.01$\\
New York Times & $+6.32$ & $+4.76$ & $-1.56$ & National Enquirer & $-4.59$ & $-1.08$ & $+3.51$\\
\bottomrule
\end{tabular}
\end{table}

\section{External Credibility Ground-Truth (MBFC)}\label{app:mbfc}
We validate the \emph{human-aligned} claim against an independent, citable professional rating: Media Bias/Fact Check \citep{mbfc2026} factual-reporting scores (lower = more factual on MBFC's scale; we correlate the model prior against credibility, i.e.\ the reverse orientation). The model's brand prior correlates with MBFC at Spearman $\rho{=}0.88$ ($p{=}0.004$), Pearson $r{=}0.89$ ($p{=}0.003$), and $\rho{=}0.82$ against the coarser MBFC credibility tier. The bootstrap CI is wide ($[0.26,1.00]$) because there are only eight brands; the permutation $p$ is the load-bearing statistic. The relationship is deliberately imperfect: the largest residual is the \emph{Daily Mail}, which MBFC rates Low/Questionable but the model scores mildly credible ($+1.48$), a concrete miscalibration consistent with our framing.
\begin{table}[h]\centering\small
\caption{Model brand prior vs.\ MBFC factual-reporting ratings \citep{mbfc2026} (accessed June 2026).}
\begin{tabular}{lccc}
\toprule
Brand & MBFC factual (lower=better) & MBFC tier & model cred\\
\midrule
Reuters & $0.0$ (Very High) & HIGH & $+5.80$\\
The New York Times & $1.4$ (High) & HIGH & $+6.32$\\
The Guardian & $1.8$ (High) & HIGH & $+4.42$\\
BBC & $2.1$ (High) & HIGH & $+3.91$\\
The Sun & $5.0$ (Mixed/Low) & LOW & $-0.66$\\
Daily Mirror & $6.0$ (Low) & LOW & $+1.06$\\
Daily Mail & $7.1$ (Low) & LOW & $+1.48$\\
National Enquirer & $8.0$ (Very Low) & LOW & $-4.59$\\
\bottomrule
\end{tabular}
\end{table}
\begin{figure}[h]
\centering
\includegraphics[width=0.6\linewidth]{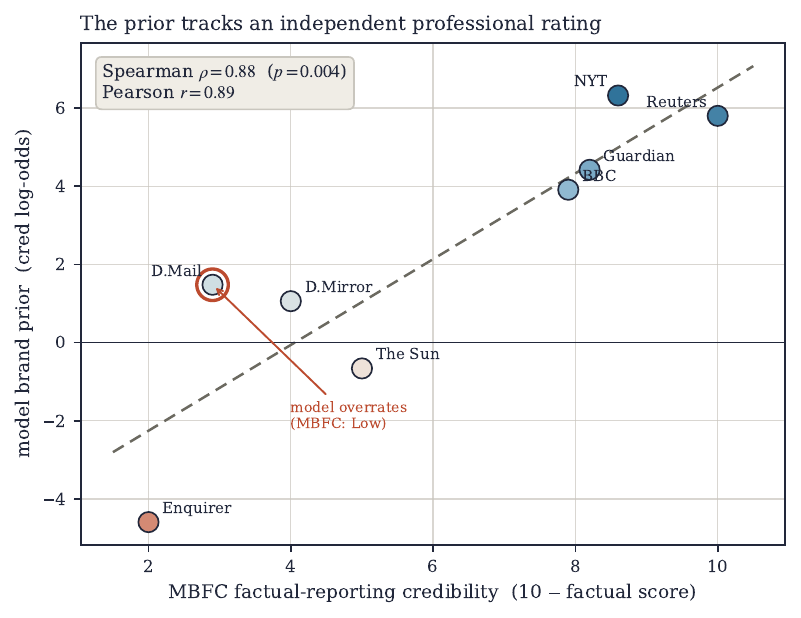}
\caption{Model brand prior versus MBFC factual-reporting credibility (Spearman $\rho{=}0.88$, $p{=}0.004$); the \emph{Daily Mail} (circled) is the largest residual, rated Low by MBFC yet mildly credible by the model.}
\label{fig:mbfcscatter}
\end{figure}

MBFC values follow the version accessed June 2026 and should be re-pinned to a dated snapshot before publication.

\section{Models, Environment, and Statistics}
\begin{table}[h]\centering\small
\caption{Scaling across seven models and two families (brand-name channel): credibility range and Spearman agreement with the human order $\rho$; model size is in the name.}
\label{tab:scaling}
\begin{tabular}{lcc@{\hskip 2em}lcc}
\toprule
Model & range & $\rho$ & Model & range & $\rho$\\
\midrule
SmolVLM-2B & $0.77$ & $0.74$ & InternVL3-8B & $4.62$ & $0.91$\\
InternVL3-2B & $2.17$ & $0.71$ & Qwen2.5-VL-7B & $10.90$ & $1.00$\\
Qwen2.5-VL-3B & $3.87$ & $0.95$ & InternVL3-14B & $12.89$ & $0.91$\\
LLaVA-OV-7B & $2.27$ & $0.76$ & & &\\
\bottomrule
\end{tabular}
\end{table}
Primary model Qwen2.5-VL-7B \citep{bai2025qwen25vl} (4-bit) on a single 24\,GB GPU. Scaling set: SmolVLM-2B \citep{marafioti2025smolvlm}, InternVL3-2B/8B/14B \citep{zhu2025internvl3}, Qwen2.5-VL-3B/7B, LLaVA-OV-7B \citep{li2024llavaonevision}. Cross-family mechanism: InternVL3-8B (28 layers). Gemma-3-4B was excluded from scaling: it requires bf16 (4-bit gives NaN via attention soft-capping) and its brand ranking is scrambled, making it non-comparable. Raw cred logits are not comparable across models (logit scales differ), so cross-model analysis uses within-model range, high$-$low gap, and Spearman vs human order. Statistics: bootstrap $n{=}5000$ for CIs, permutation $n{=}10000$ for significance.

\section{Held-out-brand Generalization of the Intervention}\label{app:heldout}
We render seven outlets absent from the main set (high: NPR, Associated Press, Al Jazeera, PBS NewsHour; low: Breitbart, InfoWars, Daily Star) as brand-name wordmarks over the six articles and steer the \emph{unchanged} $-$credibility direction (feature selected on the original eight brands) into the held-out high brands. The held-out prior is itself strong, high$-$low gap $+9.33$ (permutation $p=5\times10^{-4}$), with InfoWars $-11.1$ and Breitbart $-9.2$ below the original tabloids. Steering moves the held-out high brands at a causal slope of $-0.039$ cred per unit $\alpha$ versus $-0.063$ in-sample, i.e.\ $61\%$ of the original slope. The direction selected on the training brands therefore transfers to unseen ones, confirming a general brand-credibility axis rather than an eight-brand artifact. (The held-out gap does not reach zero within the swept $\alpha$ because we steer only the high side and the held-out low brands are far more extreme than the original tabloids; the generalization claim rests on the causal slope, not gap closure.)

\section{Prompt-based Mitigation Baseline}\label{app:promptmit}
As a baseline for the steering intervention we recompute the source-override under debiasing prompts. A strong instruction (\emph{weigh only the evidence in the article, ignore which outlet published it}) collapses the override ratio from $1.79$ to $0.05$ (override gap $+5.79\rightarrow+0.65$, $-89\%$; conflict cell $+1.58\rightarrow-6.67$) and \emph{raises} within-brand content sensitivity ($3.29\rightarrow8.02$). The prompt thus suppresses the override more than the latent intervention ($-41\%$, \S\ref{sec:override}). We report this plainly: our claim for the steering direction is interpretability, continuous and selective control, and independence from instruction-following, not maximal suppression. The prompt result also reinforces the central dissociation, the model \emph{can} weight content when instructed, so the default-inference override is a deployment gap rather than a capability deficit; whether prompting acts \emph{through} the same L20 direction is left to future work.

\begin{figure}[t]
\centering
\includegraphics[width=0.85\linewidth]{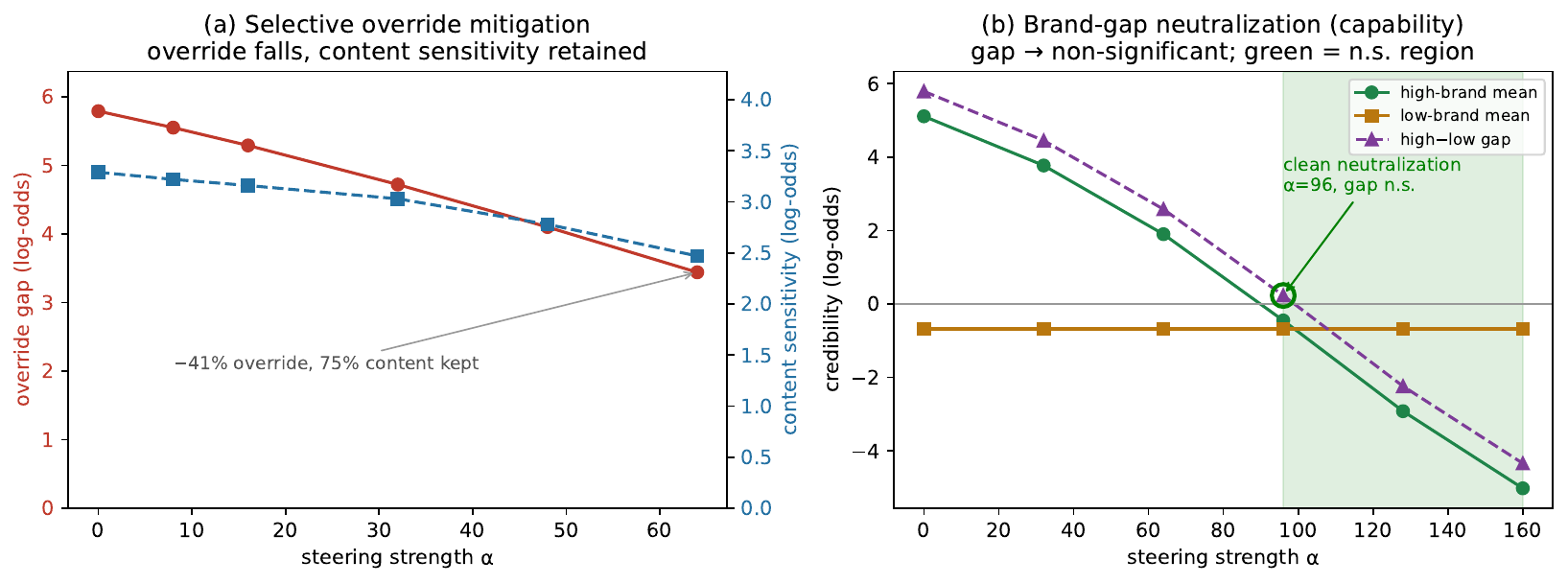}
\caption{The override is causally addressable through the identified direction. (a) Steering the $-$credibility direction into high-brand stimuli reduces the override gap by $41\%$ while retaining $75\%$ of content sensitivity (selective: the low-brand control cell is unchanged). (b) Pushed further, the same direction neutralizes the high$-$low brand gap to non-significance at $\alpha{=}96$ (shaded) without inversion, a capability demonstration.}
\label{fig:tradeoff}
\end{figure}
\section{CueTrust Benchmark: Construction, Protocol, and Statistics}\label{app:cuetrust}
\smallskip\noindent\textbf{Task and cues.} Each CueTrust cell is a $2{\times}2$ design crossing a source cue (high vs.\ low credibility) with the article's content evidence (well-sourced vs.\ red-flag), rendered over the six neutral articles (\S\ref{sec:setup}); the model is asked the single credibility question of Eq.~\ref{eq:cred}. For every non-brand cue the outlet is held constant (a generic masthead, no logo), so only the cue under test varies. The five cues are: \emph{Brand}, the masthead of a high- vs.\ low-credibility outlet (the eight outlets of \S\ref{sec:phenom}), shown as a name string or a logo image; \emph{Domain}, a browser address bar reading \texttt{nytimes.com} vs.\ \texttt{infowars.com} on a generic-masthead page; \emph{Author}, a byline naming a well-known journalist (\emph{``By Bob Woodward''}) vs.\ \emph{``By a contributor''}; \emph{Authority}, an in-text attribution to a named credentialed expert who reviewed public records vs.\ an anonymous \emph{``person familiar with the matter''}; and \emph{Format}, a broadsheet layout vs.\ a salient X/Twitter-style social-media layout (platform bar, verified badge, engagement counts), brand held constant.

\smallskip\noindent\textbf{Design rationale.} The cues separate \emph{outlet identity} from every nearby confound. Three encode the publishing outlet's identity in different surface forms (name, logo, domain); three are plausible authoritative-looking cues that are \emph{not} outlet identity (a credentialed author, an in-text expert attribution, a professional layout) and serve as negative controls. This directly answers the central confound a reviewer would raise, that the model merely reacts to any authority signal: were that so, author, authority, and format would also override; they do not. We initially hypothesized a broader \emph{presentation-level vs.\ content-embedded} boundary (any cue in the page's visual presentation overrides, any cue inside the article text does not); the format null, a presentation-level cue that does not override, falsifies that broader claim and sharpens it to outlet identity specifically. The $2{\times}2$ structure isolates the cue effect from the content effect, and the Source-Override Index (Eq.~\ref{eq:soi}) reports their ratio, a within-model quantity comparable across models whose raw credibility logits are on different scales.

\smallskip\noindent\textbf{Computation.} For each (cue, model) we fit the OLS regression of Algorithm~\ref{alg:soi}, $\mathrm{cred}\sim\beta_0+\beta_{\text{cue}}\mathbf{1}[\text{high}]+\beta_{\text{content}}\mathbf{1}[\text{well-sourced}]$, and report $\mathrm{SOI}=|\beta_{\text{cue}}|/|\beta_{\text{content}}|$. The raw $\beta$'s are not comparable across models, but their ratio is. The matrix (Table~\ref{tab:cuetrust}) reuses the rendered stimuli across all seven models, so cross-model differences reflect the models, not the inputs.

\smallskip\noindent\textbf{Statistics.} Significance of each cue is a permutation test on the cue main effect (high$-$low), shuffling the cue label $2000$ times. Where $\mathrm{SOI}>1$ the effect is significant (brand and domain $p\le5\times10^{-4}$); the negative-control cues are non-significant across models (author, authority, and format all $p>0.05$), with two isolated exceptions (LLaVA format $p=0.003$; InternVL author $p=0.026$) that do not reach $\mathrm{SOI}>1$. Content is controlled: the well-sourced and red-flag bodies carry identical facts and the same named source, are length-matched within $8\%$, and share body font size and line height across skins, so a cell's content axis differs only in epistemic markers.

\smallskip\noindent\textbf{Generation pipeline.} Stimuli use the same HTML$\rightarrow$PNG renderer as the main set (\S\ref{sec:setup}, $1024{\times}768$), with a per-cue generator injecting the manipulation into a shared template, the browser/domain chrome (domain), the byline (author), an in-text attribution sentence (authority), or the page skin (format), holding all other elements fixed. Each cue's set is rendered once and scored on every model; generators, frozen content hashes, and the SOI/permutation code are released with the stimulus suite.

\section{Formal Procedures}\label{app:algorithms}
\begin{algorithm}[h]
\caption{CueTrust: Source-Override Index (SOI) for one cue on one model.}
\label{alg:soi}
\begin{algorithmic}[1]
\Require model $M$; stimuli $\{x\}$ labeled by cue $c(x)\in\{\text{high},\text{low}\}$ and content $q(x)\in\{\text{ws},\text{rf}\}$
\For{each $x$}
  \State $s(x)\gets z_M(\text{`` Yes''}\mid x)-z_M(\text{`` No''}\mid x)$ \Comment{credibility logit, Eq.~\ref{eq:cred}}
\EndFor
\State fit OLS $\; s \sim \beta_0+\beta_{\text{cue}}\,\mathbf{1}[c(x){=}\text{high}]+\beta_{\text{content}}\,\mathbf{1}[q(x){=}\text{ws}]$
\State $\mathrm{SOI}\gets |\beta_{\text{cue}}|/|\beta_{\text{content}}|$ \Comment{Eq.~\ref{eq:soi}; within-model, comparable across models}
\State $\Delta\gets \operatorname{mean}_{c=\text{high}} s-\operatorname{mean}_{c=\text{low}} s$;\; $p\gets$ permutation test on $\Delta$ (shuffle cue labels, $2000$ draws)
\State \Return $\mathrm{SOI},\,\beta_{\text{cue}},\,\beta_{\text{content}},\,p$ \Comment{one CueTrust cell; sweep cues$\times$models for the matrix}
\end{algorithmic}
\end{algorithm}

\begin{algorithm}[h]
\caption{Localization $\rightarrow$ intervention (mechanism and method).}
\label{alg:tda}
\begin{algorithmic}[1]
\Require model $M$; matched stimulus pairs $(x,x')$ differing only in the source cue
\For{each layer $l$}
  \State $R^2(l)\gets \mathrm{CV\text{-}ridge}\big(h^{(l)}(x)\to s(x)\big)$ \Comment{linear probe with control-task selectivity}
\EndFor
\State $l^\ast\gets \arg\max_l R^2(l)$ \Comment{candidate locus}
\For{each layer $l$}
  \State $T(l)\gets \big[s(x;\,h^{(l)}\!\gets\!h^{(l)}(x'))-s(x)\big]\big/\big[s(x')-s(x)\big]$ \Comment{activation patching, Eq.~\ref{eq:transfer}}
\EndFor
\State confirm $T(l)\!\to\!1$ around $l^\ast$ \Comment{causal locus, L19--21}
\State train a TopK SAE at $l^\ast$; $d_{\text{neg}}\gets$ decoder direction of the credibility-negative feature
\For{$\alpha$ in sweep}
  \State $h^{(l^\ast)}[P]\gets h^{(l^\ast)}[P]+\alpha\,d_{\text{neg}}$ into high-credibility sources \Comment{steering intervention}
\EndFor
\State \Return locus $l^\ast$, feature direction $d_{\text{neg}}$, and the $\alpha$-effect curve
\end{algorithmic}
\end{algorithm}

\end{document}